%% file: main.tex
\documentclass{article}
\usepackage[final]{neurips_2020}
\pdfoutput=1

\usepackage{booktabs} 
\usepackage{graphicx}
\usepackage{enumitem}
\usepackage{tabularx}
\usepackage{subcaption}
\usepackage{hyperref}
\usepackage{titlesec}
\usepackage{url}
\usepackage{listings}
\usepackage{xspace}
\usepackage{xcolor} 
\usepackage[skip=3pt]{caption}
\usepackage{xurl}
\usepackage{colortbl}
\usepackage[numbers]{natbib}

\definecolor{Brown}{cmyk}{0,0.81,1,0.60}
\definecolor{OliveGreen}{cmyk}{0.64,0,0.95,0.40}
\definecolor{CadetBlue}{cmyk}{0.62,0.57,0.23,0}

\colorlet{punct}{red!60!black}
\definecolor{background}{HTML}{EEEEEE}
\definecolor{delim}{RGB}{20,105,176}
\colorlet{numb}{magenta!60!black}

\lstdefinelanguage{json}{
    basicstyle=\normalfont\ttfamily,
    numbers=left,
    numberstyle=\scriptsize,
    stepnumber=1,
    numbersep=8pt,
    showstringspaces=false,
    breaklines=true,
    frame=lines,
    backgroundcolor=\color{background},
    literate=
     *{0}{{{\color{numb}0}}}{1}
      {1}{{{\color{numb}1}}}{1}
      {2}{{{\color{numb}2}}}{1}
      {3}{{{\color{numb}3}}}{1}
      {4}{{{\color{numb}4}}}{1}
      {5}{{{\color{numb}5}}}{1}
      {6}{{{\color{numb}6}}}{1}
      {7}{{{\color{numb}7}}}{1}
      {8}{{{\color{numb}8}}}{1}
      {9}{{{\color{numb}9}}}{1}
      {:}{{{\color{punct}{:}}}}{1}
      {,}{{{\color{punct}{,}}}}{1}
      {\{}{{{\color{delim}{\{}}}}{1}
      {\}}{{{\color{delim}{\}}}}}{1}
      {[}{{{\color{delim}{[}}}}{1}
      {]}{{{\color{delim}{]}}}}{1},
}

\newcommand{\resonance}{\textit{Resonance}\xspace}
\newcommand{\skype}{\textit{Skype}\xspace}

\titlespacing{\section}{0pt}{\parskip}{-0.5\parskip}

\title{Resonance: Replacing Software Constants with \\
    Context-Aware Models in Real-time Communication}

\author{%
Jayant Gupchup, 
Ashkan Aazami, 
Yaran Fan, 
Senja Filipi, 
Tom Finley, 
Scott Inglis,\\
{\bf 
Marcus Asteborg, 
Luke Caroll, 
Rajan Chari, 
Markus Cozowicz, 
Vishak Gopal, 
Vinod Prakash,
}\\
{\bf
Sasikanth Bendapudi, 
Jack Gerrits, 
Eric Lau, 
Huazhou Liu\thanks{Presently affiliated with Outreach.io. Author performed work while at Microsoft} , 
Marco Rossi, 
Dima Slobodianyk,
}\\
{\bf
Dmitri Birjukov, 
Matty Cooper, 
Nilesh Javar, 
Dmitriy Perednya,
}\\
{\bf
Sriram Srinivasan, 
John Langford, 
Ross Cutler, 
Johannes Gehrke
}
\\\\
Microsoft
}

\newcommand{\eat}[1]{}

\begin{document}

\maketitle


\input{abstract}

\section{Introduction}
\label{s:intro}
\input{intro}



\section{Resonance}
\label{s:system}
\input{system}

\section{Experimental Results}
\label{s:casestudy}
\input{experiments}


\section{Discussion}
\label{s:conclusion}
\input{conclusion}

\bibliographystyle{abbrv}

\end{document}

%% file: abstract.tex
\begin{abstract}


Large software systems tune hundreds of `constants' to optimize their runtime performance. These values are commonly derived through intuition, lab tests, or A/B tests. A `one-size-fits-all' approach is often sub-optimal as the best value depends on runtime context. In this paper, we provide an experimental approach to replace constants with learned contextual functions for \skype - a widely used real-time communication (RTC) application. We present \resonance, a system based on contextual bandits (CB). We describe experiences from three real-world experiments: applying it to the audio, video, and transport components in \skype. We surface a unique and practical challenge of performing machine learning (ML) inference in large software systems written using encapsulation principles. Finally, we open-source \emph{FeatureBroker}, a library to reduce the friction in adopting ML models in such development environments.

\end{abstract}

%% file: intro.tex
Hyperparameter tuning of ML models using automation has a very rich literature \cite{golovin2017google, akiba2019optuna, perrone2019learning, elsken2019neural, jin2019auto, paul2019fast}. However, tuning of application constants such as buffer sizes, thresholds, or timeout settings is commonly done manually. Tuning these constants in live production systems requires overcoming many practical hurdles. In this paper, we study these challenges for a widely used RTC application. 

\begin{figure}[t]
  \begin{subfigure}[t]{0.35\linewidth}
  \includegraphics[width=0.7\linewidth]{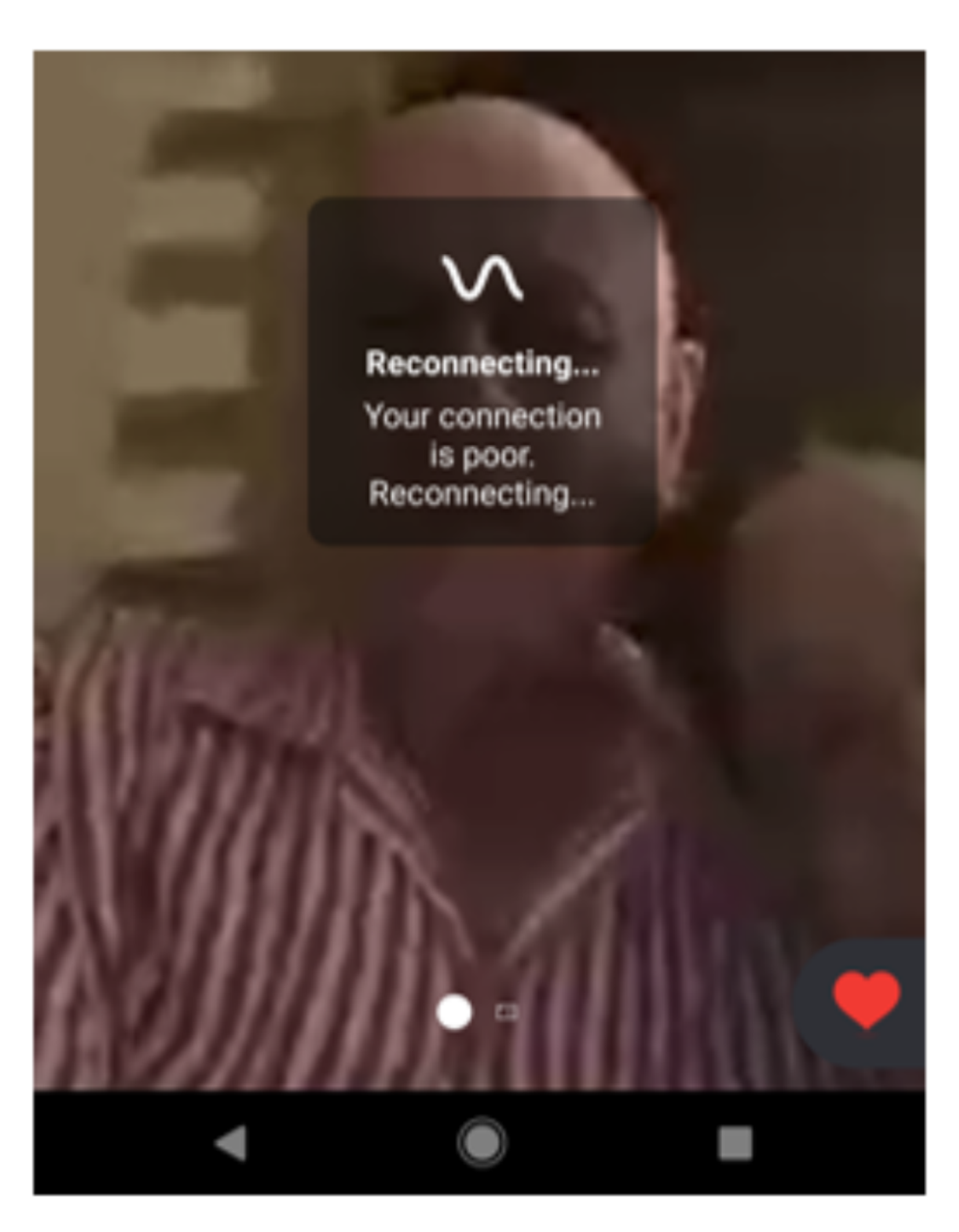}
  \label{fig:Ng1} 
  \end{subfigure}
  \begin{subfigure}[t]{0.60\linewidth}
  \includegraphics[width=0.9\linewidth]{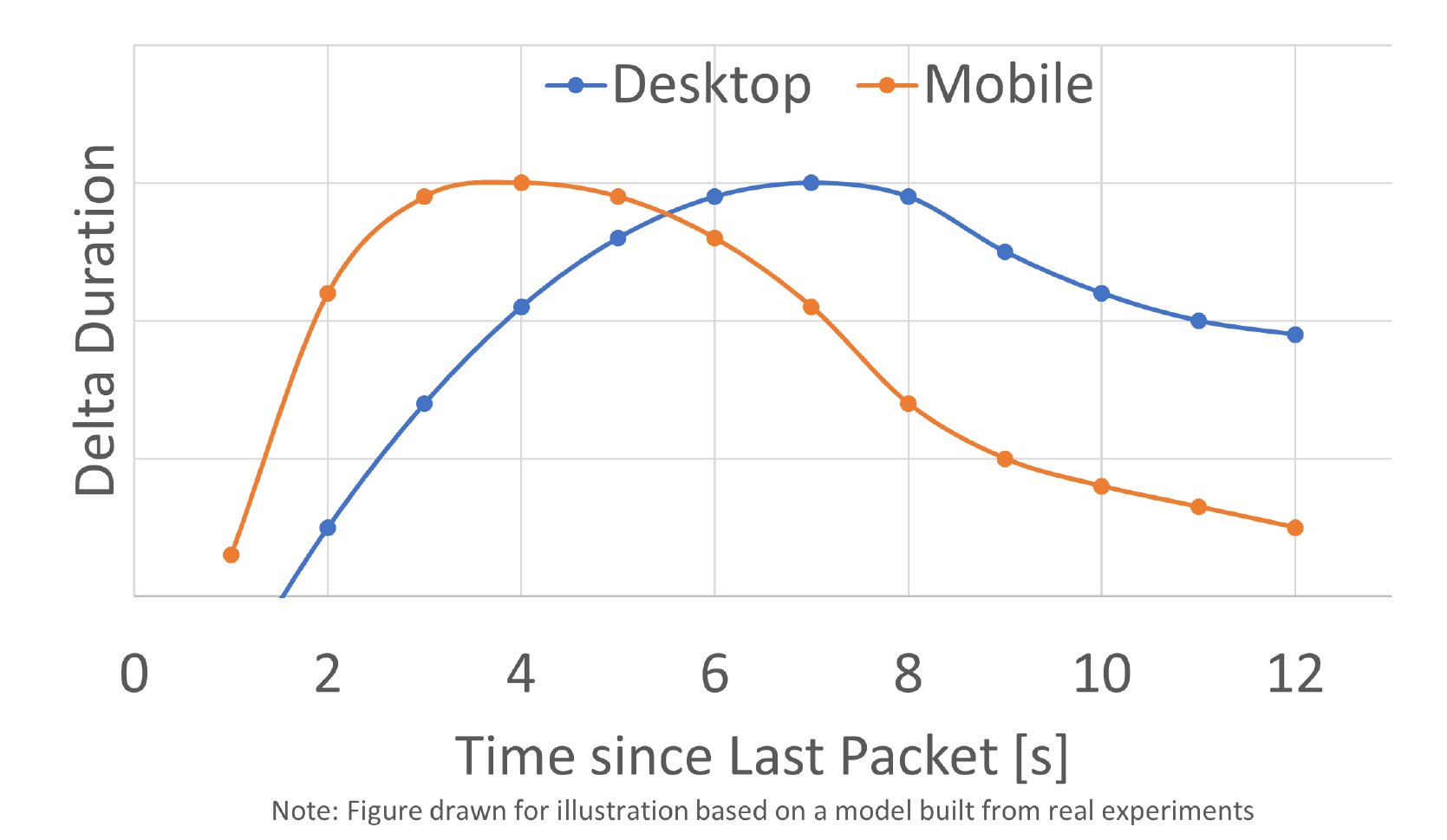}
  \label{fig:Ng2}
  \end{subfigure}
\caption[]{The left figure shows the \emph{Network Reconnect} feature in \skype. The right figure shows the impact of trigger threshold on average call duration (ACD). \footnotemark }
\label{fig:networkreconnect}
\end{figure}

\textbf{Motivating Example}: Consider a video conference call in \skype where one of the participants is using an unstable network. In \skype, if the endpoint stops receiving media packets, the application attempts to ``reconnect'' the call on an alternate interface (e.g., WiFi interface stops working, but 4G network is active). The reconnect threshold constant represents a careful trade-off. If the switch is triggered too early, the user's connection is unnecessarily moved into the ``call reconnecting'' status and starts to renegotiate all call parameters, leading to a false positive (a disruptive experience). If the application waits too long, the user loses patience and hangs up due to frustration. Furthermore, we found mobile users tend to hang up more quickly than desktop users, indicating the contextual nature of this problem as shown in Figure \ref{fig:networkreconnect}. 


\textbf{Literature}: Recently, a few solutions and systems have emerged to address some of these challenges. Carbine et al. present \emph{SmartChoices}, an API to replace heuristics in traditional algorithms with ML models built using standard reinforcement learning (RL) methods \cite{smartchoices}. \emph{SmartChoices} demonstrated the effectiveness on search, sort (pivot index), and cache algorithms. \cite{mao2016resource} and \cite{krishnan2018learning} are examples of applying RL based methods in query optimization and job scheduling systems respectively. To replace runtime constants, we employ a simpler approach of CB as we found it to scale effectively in our production scenarios compared to RL. This observation has been echoed in \cite{bakshy2018ae, eytantalk, dulac2019challenges}. In \cite{eytantalk} Bakshy et al. describe \emph{Ax}, a CB-based optimization and experimentation system for learning parameters at Facebook. In \resonance, we address a novel and practical challenge of ML inference in real-world systems when context information is trapped behind API surfaces.

\textbf{Inference Challenges}: Large production software systems comprise of components interacting via strict API boundaries (or contracts). This encapsulation creates isolation, making it difficult to surface context (or features) to be consumed by other components for runtime inference (detailed example in Section \ref{s:system}). A naive approach requires several layers of API changes and alignment that result in the breaking of established contracts. We find that these high development costs were a big driver for the lack of adoption (or integration) of CB models relying on distributed context. It is noteworthy that this challenge is not unique to CB models, but rather any model requiring access to features distributed in siloed components. To scale adoption, we created a library called \emph{FeatureBroker} that was able to address these challenges and bring down integration costs from 2 months to 2 days.

\textbf{Contributions}: This paper makes the following contributions:
\begin{itemize}[leftmargin=*,noitemsep,topsep=0pt]
  \item We outline \emph{Resonance}, an experimental system to replace constants with CB models for RTC scenarios.
  \item We detail \emph{FeatureBroker}, a library for managing the conversation between an inference engine and distributed features produced by components. We make this library open source for the benefit of the community \cite{featurebroker}.
   \item We present results from applying \emph{FeatureBroker} to optimize \skype's audio, video, and transport components in production (Section \ref{s:casestudy}).
\end{itemize}

\setlength{\belowcaptionskip}{-7pt}
\footnotetext{Figure drawn for illustration by modeling data gathered from real telemetry.}

%% file: system.tex
\textbf{System Overview}: \resonance comprises of the following components: \emph{Skype App}, \emph{A/B test system}, \emph{Client Inference (CI)} and \emph{Learning Service (LS)} as described in \cite{agarwal2016multiworld}. The benefit of replacing an \emph{App} constant is evaluated as an A/B test; the control variant represents the fixed constant, whereas the treatment variant is represented by a CB model built using \emph{VowpalWabbit (VW)} \cite{langford2007vowpal} -- a well known open source library for CB models. The \emph{CI} component performs runtime inference based on the \emph{App} context as specified by the model. The \emph{CI} component captures training logs comprising of \{context value, recommended action, action probability, reward metric\} and transmits them to the \emph{LS}. The \emph{LS} uses these training logs to produce a \emph{VW} models (or policies) that get delivered to the client endpoints periodically (e.g., every 4 hours) completing the inference-log-learn loop. 

In this paper, we focus our attention on the novel \emph{FeatureBroker} client library. In our explanations, we define context as a collection of feature values (e.g., network type = WiFi, platform = mobile).

\setlength{\belowcaptionskip}{-7pt}
\begin{figure}[t]
    \begin{subfigure}{0.42\linewidth}
      \includegraphics[width=0.95\linewidth]{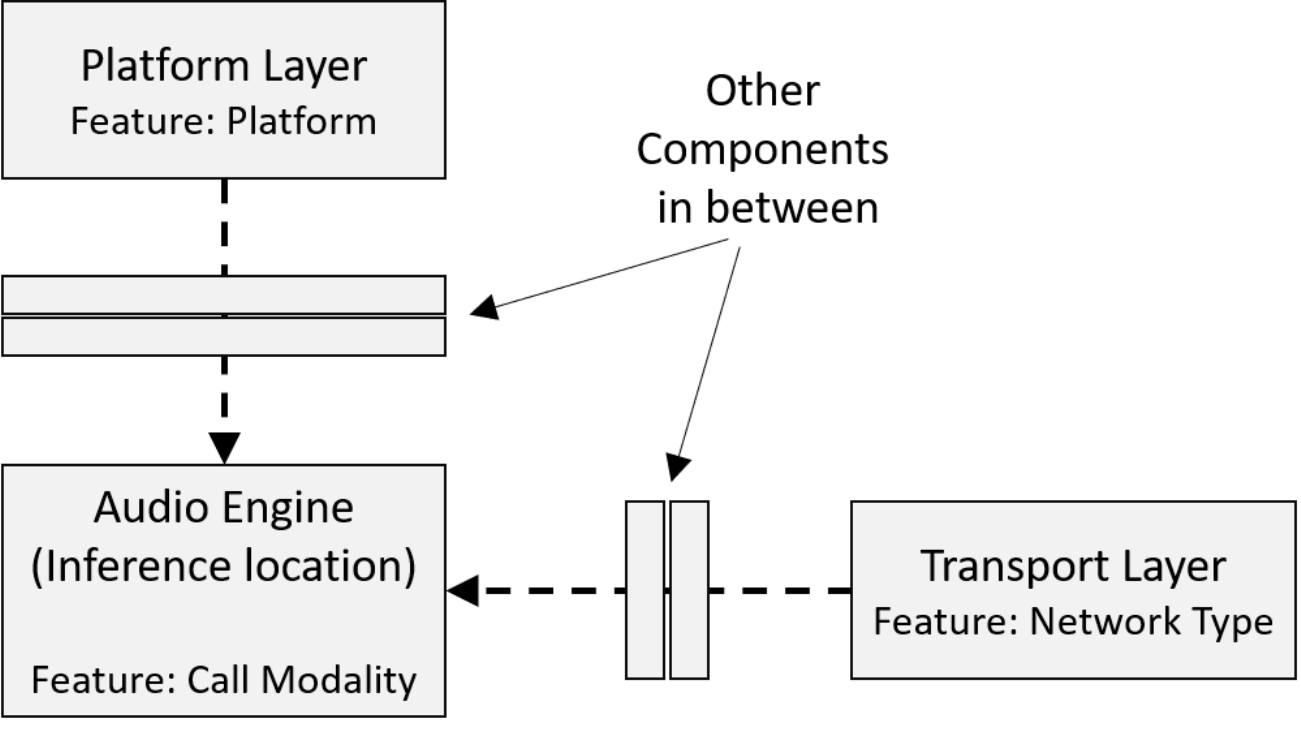}
      \label{fig:fb_components_audio}
     \end{subfigure}
    \begin{subfigure}{0.53\linewidth}
      \begin{lstlisting}[gobble=8, basicstyle=\footnotesize, language=C++, numbers=none,
        caption={}]
        // Inference without FeatureBroker
        jbHysteresis = model.predict([
            FeatureStore.GetValue("nwType"),
            FeatureStore.GetValue("platform"),
            FeatureStore.GetValue("callType")]);
        \end{lstlisting}
    \end{subfigure}
      
   \caption{Left figure shows the organization of components within 
   \skype's RTC stack; illustrating the distributed nature of features in a well-encapsulated system. The right figure shows the inference performed in the absence of a library like \emph{FeatureBroker}.}
   \label{fig:fb}
\end{figure}

\textbf{Feature Broker}: We motivate the need for such a library by way of an example. Consider we want to replace a constant \emph{jbHysteris} in the audio component with the output of a CB model. A simple API to do that is shown in Figure \ref{fig:fb} (right side).

While this API looks simple, practical problems
emerge when performing inference using this approach. In this scenario, the context depends on the global state of the system, such as the platform (e.g.,
desktop/wired), the network type (e.g., wired/WiFi), and call modality
(e.g., audio/video). Figure \ref{fig:fb} (left side) shows the logical organization of components within our RTC application. While call modality information is available within the audio component, platform and network type
information were produced in different components. Moreover, the audio and platform/transport components were separated by many other components and API surfaces. This situation where the features of interest are encapsulated in well-separated components was found to be a common pattern. Component owners want well-defined, stable public API surfaces for maintenance reasons. In addition, expanding them on a feature-by-feature basis is onerous. Furthermore, the components producing the features may not be running in the same thread as the component performing inference, and lastly, the changes in feature values at runtime (e.g., a mid-call change in network type) would require updating the prediction gracefully. 

The idea of a simple shared thread-safe key-value feature store proved insufficient for two primary
reasons:

\begin{enumerate}[leftmargin=*,noitemsep,topsep=-5pt]
    \item \emph{Coherence}:  Inference libraries need feature values to be
     stable and consistent. However, as seen
    above, feature values can update asynchronously (e.g., network type can change mid-call).
    \item \emph{Hierarchy}: Features may be consumed by multiple models and inference components. For example, three models (\emph{JBHysteresis},
    \emph{ScreenshareEncoding} and \emph{NetworkReconnect}) may consume platform
    and network type features while having their own local features
    (e.g., call modality). The most graceful way to handle this was to make the
    structure hierarchical and allow it to be ``forked'' for sub-components.
\end{enumerate}

The shared structure responsible for handling these problems is termed \emph{FeatureBroker}. It manages the conversation between the client code that provides features (inputs) and consumes inference results
(outputs) from a CB model. In this paper, it would be impossible to describe the library in detail. Instead, we highlight the main elements of the solution. Interested readers can see a detailed real-world example in the open-source version of the library \emph{FeatureBroker} \cite{featurebroker}. The key operations of the \emph{FeatureBroker} are:

\noindent \begin{enumerate}[leftmargin=*,noitemsep,topsep=-2pt]
    \item \emph{Binding Inputs}: Given a name and type returns an ``input pipe'' into which a feature providing component can feed values.
    \item \emph{Associating Models}: Register and describe a scheme for transforming inputs to outputs.
    \item \emph{Binding Outputs}: Returns an ``output pipe'' in which inference consuming models can query for updated values. 
\end{enumerate}

The notion of binding draws parallels to binding values for form elements obtained from a database \cite{lerman2010programming}. Finally, it is noteworthy that \emph{FeatureBroker} can be used with any inference library (e.g., ONNX \cite{onnxruntime}) and the concept even generalizes beyond CB models - the problem is most amplified in CB inference due to the distributed nature of context generation. As stated earlier, detailed examples are presented in \cite{featurebroker}.




\eat{

    \textbf{\emph{Slim VW} and \emph{FeatureBroker}}: The \emph{FeatureBroker} and \emph{Slim VW}
    components are responsible for performing inference within the running process.
    \emph{Slim VW} is the inference portion of the popular open source contextual
    bandit library VowpalWabbit (VW) \cite{langford2007vowpal}. As part of this
    work, the C++ version of \emph{slim VW} was made available to the open source
    community \cite{slimvw}. The \emph{FeatureBroker} library works in conjunction
    with \emph{slim VW} and manages the conversation between features and inference.
    The design details of the \emph{FeatureBroker} will be described in further
    detail in Section \ref{subs:fb}.

    \textbf{Event Processing System}: The inference and rewards telemetry (or logs)
    are uploaded by a client library and processed by the real-time group's data
    pipeline system. The data pipeline system is built using Microsoft's stream
    processing system known as \emph{Siphon} \cite{siphon}. Commercial alternatives
    to this component are also available. 

    \textbf{Resonance Service}: The training and reward logs are forwarded to the
    \resonance service where these logs are processed, shaped and forwarded to
    the Azure Personalization Service \cite{aps}. In addition to the training and reward
    processing, the resonance service is responsible for authentication, compliance
    management and retrieving a trained model on the schedule requested by the
    experimenter. The design details of the \emph{Resonance service} will be described in further
    detail in Section \ref{subs:service}.

    \textbf{Azure Personalizer Service (APS)}: APS is a commercially available
    service that implements different contextual bandit (CB) algorithms for
    personalization. The \resonance system leverages APS for replacing fixed
    constants with context-aware values in an online fashion. The model snapshots
    are delivered back to the client via the experimentation system.

    \textbf{Experimentation System}: Experimenters parameterize the fixed constants
    such that the value can be set by the real-time communication's experimentation
    (A/B) system. \resonance builds on top of this infrastructure by running a
    A/B tests where the control is the fixed (or existing) policy whereas the
    treatment is the context-aware CB model. Building on top of the A/B system
    provides multiple benefits such as gradual exposure to users (safety), causality
    of results and rich tooling of A/B testing (such as sample ratio mismatch
    \cite{fabijan2019diagnosing}).

    \textbf{Analysis Tools}: In addition to an hourly version of the experimentation
    scorecard, we built a tool called \emph{ContextExplorer} that provided
    experimenters with insights about the CB-based model. As part of this work, we
    have open sourced \emph{ContextExplorer} for benefitting the community that
    will be using \emph{VW} or \emph{APS}. We will cover \emph{ContextExplorer} in
    more depth in Section \ref{subs:analysis}.

    We provide further detail on the new infrastructure that has been built for
    \resonance.

    \subsection{Resonance Service}
    \label{subs:service}

\input{service}

}

%% file: service.tex
In recommender systems, inference and training happens in a backend service like
APS. In our RTC application, we don't have the luxury of waiting for a service
call to complete . In \resonance, the inference is done on the client
endpoint using (i.e., \emph{slim VW}, \cite{slimvw}). The training is done in a
batched fashion using APS by setting the \emph{VW} hyperparameters in the
personalizer instance. Each experiment maps to one personalizer instance.

A natural question to ask is: If the training is done using APS, why do we need
the \resonance service at all? We outline the design goals and
responsibilities of the \resonance service below:

\textbf{Security, Privacy and Compliance}: \resonance implements the
security, privacy and compliance policies outlined by Microsoft's RTC group to
ensure that the data handling is done within a trusted boundary. The telemetry
logs are authenticated and validated to protect against any attack vectors.
\resonance service performs telemetry scrubbing to ensure that there is no
user content, user identifying, or even pseudo identifying (hashed user id)
information in the inference logs. The inference logs forwarded to APS only
contain system metadata. \resonance service implements compliance
requirements around GDPR (geographic region handling) and data retention
policies. These requirements are particularly strict in enterprise VoIP
scenarios. Implementing all organizational data processing policies is beyond
the scope of a generic service like APS so we need to front it with a service
that implements those policies.

\textbf{Compression of logs}: The inference and reward logs need to be submitted
by client apps under various bandwidth conditions. In several countries
bandwidth is so low that it is critical to compress the logs to minimize data
loss \cite{gupchup2018trustworthy}. We employ a custom bit-packing compression
scheme to minimize log size. The \resonance service decompresses these
logs and makes use of the APS API for the purpose of registering training
records. The design of the bit-packing scheme is out of the scope of this paper.

\textbf{Reward Processing}: \resonance is built on top of \emph{Skype's}
experimentation system. The reward metric is obtained from the experimentation
pipeline. The system needs to correlate the logs and rewards within this
pipeline requiring custom event handling logic. The service also keeps track of
the the training events received, lost, corrupted and incomplete (missing
reward). It surfaces this information to experimenters via a dashboard.

\textbf{Experiment Management}: The \resonance system is designed to scale
to 100s of experiments running in parallel. It performs the role of managing a
pool of personalizer instances, forwarding training data gathered from an
experiment to the correct personalizer instance. Lastly, it manages model
promotion and delivery back to the client. The delivery cadence is set during
the experiment registration time. The service is responsible for honoring the
schedule. Larger models are stored on a CDN that can be downloaded by the
client. The link to the CDN is updating using the experimentation APIs. This
authenticated update is handled by the \resonance service.

%% file: experiments.tex
\setlength{\belowcaptionskip}{-2pt}
\begin{table}[t]
    \begin{tabular}{l|ccc}
        \toprule
        \multicolumn{1}{l|}{Experiment Name} &
          \begin{tabular}[c]{@{}c@{}} \emph{Audio Jitter Buffer}\end{tabular} &     \begin{tabular}[c]{@{}c@{}}\emph{Screen Share Encoding}\end{tabular} &
          \begin{tabular}[c]{@{}c@{}}\emph{Network Reconnect} \end{tabular} \\ 
        \midrule
        Platforms & Desktop & \begin{tabular}[c]{@{}c@{}}\{Desktop, Mobile\}\end{tabular} & \begin{tabular}[c]{@{}c@{}}\{Desktop, Mobile\}\end{tabular} \\
        \hline
          \begin{tabular}[c]{@{}l@{}}Impressions (Millions) \end{tabular}  & 4M & 7.6M & 19M \\
        \hline
        \begin{tabular}[c]{@{}l@{}}RTC Component \end{tabular}  & Audio & Video & Transport \\
        \hline 
        Number of Actions                                                                 & 10        & 10        & 5          \\
        \hline
        Unique Contexts                                                                   & 14        & 440       & 390        \\
        \hline
        Metric &
          \begin{tabular}[c]{@{}c@{}} \small{Poor-Audio-rating} \end{tabular} &
          \begin{tabular}[c]{@{}c@{}} \small{Poor-Video-rating} \end{tabular} &
          \begin{tabular}[c]{@{}c@{}} \small{Call Duration (CD)} \end{tabular} \\
        \hline
        Model Size                                                                        & 4 KB      & 22 KB     & 10 KB      \\
        \hline
        \begin{tabular}[c]{@{}l@{}} Metric Improvement (\%)\end{tabular} & 1.1\%     & 9.9\%     & 4.2\%      \\
        \bottomrule
    \end{tabular}

    \vspace*{0.3 cm}


    
    \caption{The table captures the overall Summary results from experiments conducted on the representative scenarios. 
    }
    \label{Tab:all_summary}
    \vspace{-8mm}
\end{table}

We present results from three scenarios in \skype. The scenario related to learning the reconnect threshold was described in Section \ref{s:intro}. We introduce two other scenarios.

\noindent \textbf{Scenario: Audio Jitter Buffer (JB) Hysteresis} The JB component absorbs variability in network packet arrivals (i.e., jitter) to present a smooth playout of audio frames transmitted via the network. \emph{JBHysteresis} is a meta-parameter controlling the amount of inertia associated with changes in the buffer size. The optimization metric to learn this parameter was \textit{Poor-Audio-Rating}, an objective metric estimating speech quality based on packet loss, jitter, and conversational delay derived from user ratings \cite{ding2003speech}.

\noindent \textbf{Scenario: Video Encoding Bitrate Allocation for Screen Share} Screen sharing sessions in VoIP require careful selection of the bitrate. Large bit rates lead to a higher image quality, but are also susceptible to frame freezes due to packet losses under poor network conditions. 
This quality-distortion tradeoff is characterized by a constant that weights transmission rates and frame freezes. Similar to audio, we optimized for a video technical metric termed \emph{Poor-Video-Rating}.

\textbf{Experiments and Results}:
A summary of the experimental data for all three scenarios is presented in Table
~\ref{Tab:all_summary}. The data is collected using an $\epsilon$-greedy policy.
The $\epsilon$ was set to 0.2 for the experiments. For each experiment,
we report the number of unique contexts to convey the range of inputs going into the \emph{FeatureBroker}. Each of these experiments showed statistically significant improvements between the fixed-constant (default) policy and the model-based policy with a p-value less than $0.01$. We note the following:
\begin{itemize}[leftmargin=*,noitemsep,topsep=1pt]
    \item These experiments show that we could successfully replace constants with CB models for RTC scenarios. Since these experiments were done in three different components by three different teams, it demonstrates the generality of the \emph{FeatureBroker} solution.
    \item New experiments were able to build and re-use the on-boarding of features done by previous experiments. For example, the on-boarding of features for \emph{ScreenShareEncoding} experiment simplified the integration of the \emph{NetworkReconnect} model as the two models shared features. Prior to the introduction of \emph{FeatureBroker}, engineering teams would have to coordinate and align on API changes to expose features. This process would often take multiple days, and has been reduced by an order of magnitude.  
    \item Due to the limited memory and CPU budget in RTC applications, we focused on keeping the footprint of CB models small. The main reason was to scale this methodology so that it can be applied to 100s of scenarios within \skype.
\end{itemize}


%% file: conclusion.tex
This paper introduces \resonance, a system for replacing application constants with context-aware models. We built and deployed this system for the
real-time production scenarios of \skype. We presented \emph{FeatureBroker}, a  library for integrating ML models relying on global context. The library presents a solution to the problem of inference when the context is distributed across strong component boundaries. We make this library available to practitioners to bring down the adoption and development costs. Using three statistically significant real-world experiments, we showed this methodology can improve system performance. Such a methodology provides a new tool for component owners to optimize their components.

\textbf{Future Work}: The problem of efficient
discovery of sensitive constants in a live system containing thousands of constants
remains challenging. 

In conclusion, we emphasize that the ideas presented in \resonance and challenges solved by \emph{FeatureBroker} are applicable to many software systems, particularly large codebases that ship with 1000s of system constants.

%% file: main.bbl
\begin{thebibliography}{10}

    \bibitem{agarwal2016multiworld}
    A.~Agarwal, S.~Bird, M.~Cozowicz, L.~Hoang, J.~Langford, S.~Lee, J.~Li,
      D.~Melamed, G.~Oshri, O.~Ribas, et~al.
    \newblock A multiworld testing decision service.
    \newblock {\em arXiv preprint arXiv:1606.03966}, 7, 2016.
    
    \bibitem{akiba2019optuna}
    T.~Akiba, S.~Sano, T.~Yanase, T.~Ohta, and M.~Koyama.
    \newblock Optuna: A next-generation hyperparameter optimization framework.
    \newblock In {\em Proceedings of the 25th ACM SIGKDD International Conference
      on Knowledge Discovery \& Data Mining}, pages 2623--2631. ACM, 2019.
    
    \bibitem{eytantalk}
    E.~Bakshy.
    \newblock Towards simplicity in machine learning for live systems with adaptive
      experimentation.
    \newblock
      \url{https://slideslive.com/38922475/towards-simplicity-in-machine-learning-for-live-systems-with-adaptive-experimentation},
      2019.
    
    \bibitem{bakshy2018ae}
    E.~Bakshy, L.~Dworkin, B.~Karrer, K.~Kashin, B.~Letham, A.~Murthy, and
      S.~Singh.
    \newblock Ae: A domain-agnostic platform for adaptive experimentation, 2018.
    
    \bibitem{smartchoices}
    V.~Carbune, T.~Coppey, A.~Daryin, T.~Deselaers, N.~Sarda, and J.~Yagnik.
    \newblock Smartchoices: Hybridizing programming and machine learning.
    \newblock In {\em Reinforcement Learning for Real Life (RL4RealLife) Workshop
      in the 36th International Conference on Machine Learning (ICML),}, 2019.
    
    \bibitem{ding2003speech}
    L.~Ding and R.~A. Goubran.
    \newblock Speech quality prediction in voip using the extended e-model.
    \newblock In {\em GLOBECOM'03. IEEE Global Telecommunications Conference (IEEE
      Cat. No. 03CH37489)}, volume~7, pages 3974--3978. IEEE, 2003.
    
    \bibitem{dulac2019challenges}
    G.~Dulac-Arnold, D.~Mankowitz, and T.~Hester.
    \newblock Challenges of real-world reinforcement learning.
    \newblock {\em arXiv preprint arXiv:1904.12901}, 2019.
    
    \bibitem{elsken2019neural}
    T.~Elsken, J.~H. Metzen, and F.~Hutter.
    \newblock Neural architecture search: A survey.
    \newblock {\em Journal of Machine Learning Research}, 20(55):1--21, 2019.
    
    \bibitem{golovin2017google}
    D.~Golovin, B.~Solnik, S.~Moitra, G.~Kochanski, J.~Karro, and D.~Sculley.
    \newblock Google vizier: A service for black-box optimization.
    \newblock In {\em Proceedings of the 23rd ACM SIGKDD International Conference
      on Knowledge Discovery and Data Mining}, pages 1487--1495. ACM, 2017.
    
    \bibitem{jin2019auto}
    H.~Jin, Q.~Song, and X.~Hu.
    \newblock Auto-keras: An efficient neural architecture search system.
    \newblock In {\em Proceedings of the 25th ACM SIGKDD International Conference
      on Knowledge Discovery \& Data Mining}, pages 1946--1956. ACM, 2019.
    
    \bibitem{krishnan2018learning}
    S.~Krishnan, Z.~Yang, K.~Goldberg, J.~Hellerstein, and I.~Stoica.
    \newblock Learning to optimize join queries with deep reinforcement learning.
    \newblock {\em arXiv preprint arXiv:1808.03196}, 2018.
    
    \bibitem{langford2007vowpal}
    J.~Langford, L.~Li, and A.~Strehl.
    \newblock Vowpal wabbit online learning project.
    \newblock \url{https://github.com/VowpalWabbit/vowpal_wabbit/wiki}, 2007.
    
    \bibitem{lerman2010programming}
    J.~Lerman.
    \newblock {\em Programming Entity Framework: Building Data Centric Apps with
      the ADO. NET Entity Framework}.
    \newblock " O'Reilly Media, Inc.", 2010.
    
    \bibitem{mao2016resource}
    H.~Mao, M.~Alizadeh, I.~Menache, and S.~Kandula.
    \newblock Resource management with deep reinforcement learning.
    \newblock In {\em Proceedings of the 15th ACM Workshop on Hot Topics in
      Networks}, pages 50--56, 2016.
    
    \bibitem{featurebroker}
    Microsoft.
    \newblock Feature broker, a library for enabling sharing of context/features
      across code components for ml inference.
    \newblock \url{https://github.com/microsoft/FeatureBroker}, 2020.
    
    \bibitem{onnxruntime}
    {ONNX Runtime: cross-platform, high performance scoring engine for ML models}.
    \newblock \url{https://github.com/microsoft/onnxruntime}, 2019.
    
    \bibitem{paul2019fast}
    S.~Paul, V.~Kurin, and S.~Whiteson.
    \newblock Fast efficient hyperparameter tuning for policy gradient methods.
    \newblock In {\em Advances in Neural Information Processing Systems}, pages
      4616--4626, 2019.
    
    \bibitem{perrone2019learning}
    V.~Perrone, H.~Shen, M.~W. Seeger, C.~Archambeau, and R.~Jenatton.
    \newblock Learning search spaces for bayesian optimization: Another view of
      hyperparameter transfer learning.
    \newblock In {\em Advances in Neural Information Processing Systems}, pages
      12771--12781, 2019.
    
\end{thebibliography}
